\documentclass{spie}

\usepackage{cite}

\usepackage{times} 
\usepackage{indentfirst} 

\usepackage{xkeyval}

\usepackage{authblk}
\usepackage[numbers]{natbib}
\usepackage{algorithm}
\usepackage{amsmath}
\usepackage{algorithmicx}
\usepackage{setspace}
\usepackage{algpseudocode}
\usepackage{indentfirst}
\usepackage{multirow}
\usepackage{float}
\usepackage{subfig}
\usepackage{graphicx}
\usepackage{xcolor}
\usepackage{todonotes}
\setcitestyle{open={[},close={]},citesep={,}}

\title{1-point RANSAC for Circular Motion Estimation in Computed Tomography (CT)}
\author{Mikhail O. Chekanov\supit{1,2}, Oleg S. Shipitko\supit{1}, Egor I. Ershov\supit{1}
  \skiplinehalf
  \normalsize 
  \supit{1}Institute for Information Transmission Problems – IITP RAS, Bol’shoy Karetnyy Pereulok 19, Moscow, Russia, 127051. \\
  \supit{2}Moscow Institute of Physics and Technology (National Research University), Institutskiy Pereulok 9, Dolgoprudny, Moscow Region, Russia, 141700. \\
}

\begin{document}

\maketitle

\begin{abstract}

This paper proposes a RANSAC-based algorithm for determining the axial rotation angle of an object from a pair of its tomographic projections.
An equation is derived for calculating the rotation angle using one correct keypoints correspondence of two tomographic projections.
The proposed algorithm consists of the following steps: keypoints detection and matching, rotation angle estimation for each correspondence, outliers filtering with the RANSAC algorithm, finally, calculation of the desired angle by minimizing the re-projection error from the remaining correspondences.
To validate the proposed method an experimental comparison against methods based on analysis of the distribution of the angles computed from all correspondences is conducted. 

\end{abstract}

\section{Introduction}

Computed tomography is widely used in various fields: medicine~\cite{kesminiene2018cancer}, precise measurements~\cite{buratti2018applications}, agriculture~\cite{mairhofer2017x}.
In tomography the mutual trajectories of the sample, detector, and probe radiation source are usually considered known, since they are determined by the targeted movement of the setup components.
The most computationally efficient tomographic reconstruction algorithms, such as the FDK~\cite{feldkamp1984practical} and algorithm proposed by A.~Katsevich~\cite{katsevich2004improved}, rely on the geometric accuracy of the instrument and the reliably known trajectories of all its parts.
However, the realized trajectory differs from the desired one for various reasons (mechanical backlash, an error in measuring the angle of rotation of an object, thermal deformations, the slope of the sample relative to the axis of rotation), which negatively affects the quality of the reconstruction.
Thus, geometric errors are one of the main sources of reconstruction errors~\cite{ferrucci2015towards}.
So, a significant effort is regularly put on the calibration of tomography systems. Geometric calibration is a laborious and expensive process, requiring the involvement of specialists with the appropriate qualifications.
Moreover, geometrical errors inherent even in a calibrated system can still have a negative effect on the quality of reconstruction, and the quality of calibration decreases with time.
Thus, the development of methods for geometric errors correction is of high practical importance for increasing the accuracy of tomographic measurements.

Meanwhile, the problem of determining the relative motion parameters of a camera is widely studied problem in robotics and computer vision. Essentially this work is an attempt to apply the existing techniques of visual odometry to refine the motion parameters of tomography setup under the condition of a more complex model of image formation (transluent world model).
In this paper we propose a method for estimating the axial rotation angle of an object under the condition that the camera is stationary (hereinafter, the terms \textit{camera} and \textit{detector} stand for digital X-ray detector).
The optical axis of the camera is perpendicular to the axis of object rotation.
Such system setup is equivalent to the case of the circular camera motion around a stationary object which is well described by the apparatus of epipolar geometry.
Further, for simplicity, we will consider this particular motion model. The assumption made in this work is that determining the axial rotation angle will allow for correction of geometric errors related to the inaccuracy of the rotation angle measurement sensors of the detector or object.
To determine the relative motion of the camera, detection and matching of keypoints on a pair of tomographic projections are performed.
Further, the rotation angle is estimated by calculating the parameters of the epipolar correspondence based on each correspondence.
The implementation of the Random Sample Consensus (RANSAC) method is used to filter the wrong matches.
The final angle estimate is calculated by minimizing the re-projection error from the correspondences remaining after outliers removal.

\section{Related work}

The problem of geometric correction in tomography has been widely studied.
The literature offers many models of the influence of geometric errors of various kinds on the quality of reconstruction~\cite{ferrucci2016evaluating, kumar2011analysis, wenig2006examination}.
There are two main ways to eliminate the influence of these errors (geometric calibration): (1) calibration based on a reference object -- an object with known geometry that allows to evaluate and correct the influence of geometric errors by means of projective geometry~\cite{dewulf2013uncertainty, hermanek2017optimized, weiss2012geometric}; (2) calibration based on reference measuring instruments by performing a direct measurement of coordinate inaccuracies in the tomographic scheme~\cite{welkenhuyzen2014investigation, bircher2018geometry}.
Existing methods of geometric calibration make it possible to partially eliminate the geometric errors.
Usually different methods are developed to eliminate a certain type of error (scaling errors~\cite{weiss2012geometric}, rotational motion measurement errors~\cite{defrise2008perturbative}, etc.).
Calibration procedures do not take into account factors arising from the difference between the conditions of calibration experiments and the experiments with real objects.
Such factors include thermal deformations arising due to the significantly longer duration of a real tomographic experiment compared with a calibration one, as well as errors caused by a change in the geometry of the test sample or its movement under its own weight during measurements.

An alternative approach to improving the result of tomographic reconstruction is the so-called \textit{online calibration}.
Methods of online calibration allow to evaluate and correct geometric errors after or directly during the tomographic experiment~\cite{yang2017direct, xu2017simultaneous, zhang2014iterative, muders2014stable, chung2018tomosynthesis}.
It can be seen that there is a gradual transition from methods with an a priori known position and shape of the calibration object (for example,~\cite{zhang2014iterative}, 2014) to methods with the simultaneous reconstruction and and geometric error correction (\cite{xu2017simultaneous}, 2017).

An important step in tomographic reconstruction methods with a simultaneous refinement of the motion parameters is the estimation of the relative camera motion using tomographic projections.
In the general case the motion of the camera in space is described by 6 parameters (3 coordinates and 3 rotation angles).
However, the motion of the detector in a tomographic setup can be approximated by circular motion.
Such an approximation, given that the radius is fixed, reduces the number of degrees of freedom to a single one -- rotation angle.
Circular motion is a special case of camera motion which is widely covered in the literature. Its features find application in areas such as camera calibration~\cite{hernandez2007silhouette}, as well as objects 3D reconstruction from the series of images (structure from motion)~\cite{mendoncca2000camera, wong2001structure, vianello2018robust}.

To match points between pairs of tomographic projections we use the keypoints -- computational algorithms that allow to detect and describe regions of images that are invariant to various image transformations.
When matching keypoints, false correspondences inevitably arise, and consequently the angle calculated based on the wrong correspondence will be incorrect.
An additional complication in the context of X-ray images is the translucency of the object: it is difficult to determine from a single image where the point of the object is located (on the front side, on the back or inside). Moreover, analyzing the sequence of images the points diametrically opposite relative to the rotation axis move in opposite directions.
Thus, for a robust estimation of the axial rotation angle of an object methods for filtering false correspondences are necessary.

The Random Sample Consensus method is a common approach to the robust estimation of model parameters from data containing outliers~\cite{ovchinkin2018algorithm, kunina2019matching, tropin2019method}.
A number of works are known where the RANSAC method, which builds a hypothesis based on a single correspondence of points in successive images, is used to evaluate camera motion.
In a series of works~\cite{scaramuzza2009real, scaramuzza20111, scaramuzza2011performance} D.~Scaramuzza and co-authors solve the problem of visual odometry by analyzing the sequence of images taken by a camera mounted on a car.
The authors showed that single-point parameterization leads to the fastest RANSAC implementation, and also demonstrated that in the problem of a car relative motion estimation, a single-point implementation can provide accuracy comparable, and sometimes superior to the standard five-point algorithm~\cite{nister2004efficient}.
A similar problem was solved by another group of authors.
In the work~\cite{van20131} N.~V.~Cuong and co-authors confirmed the findings of D.~Scaramuzza and also presented a system using the algorithm using bundle adjustment to refine the relative motion estimation.
The authors of~\cite{civera20091, civera20101} proposed an algorithm combining RANSAC with an extended Kalman filter ~\cite{grewal2011kalman}.
The use of the Kalman filter allows reducing the parameterization of the RANSAC algorithm to one-pint by using a priori probabilistic information on camera motion.

\section{Camera motion model}

\label{sec:cam_motion}
\begin{figure}
    \centering
        \includegraphics[width=0.6\textwidth]{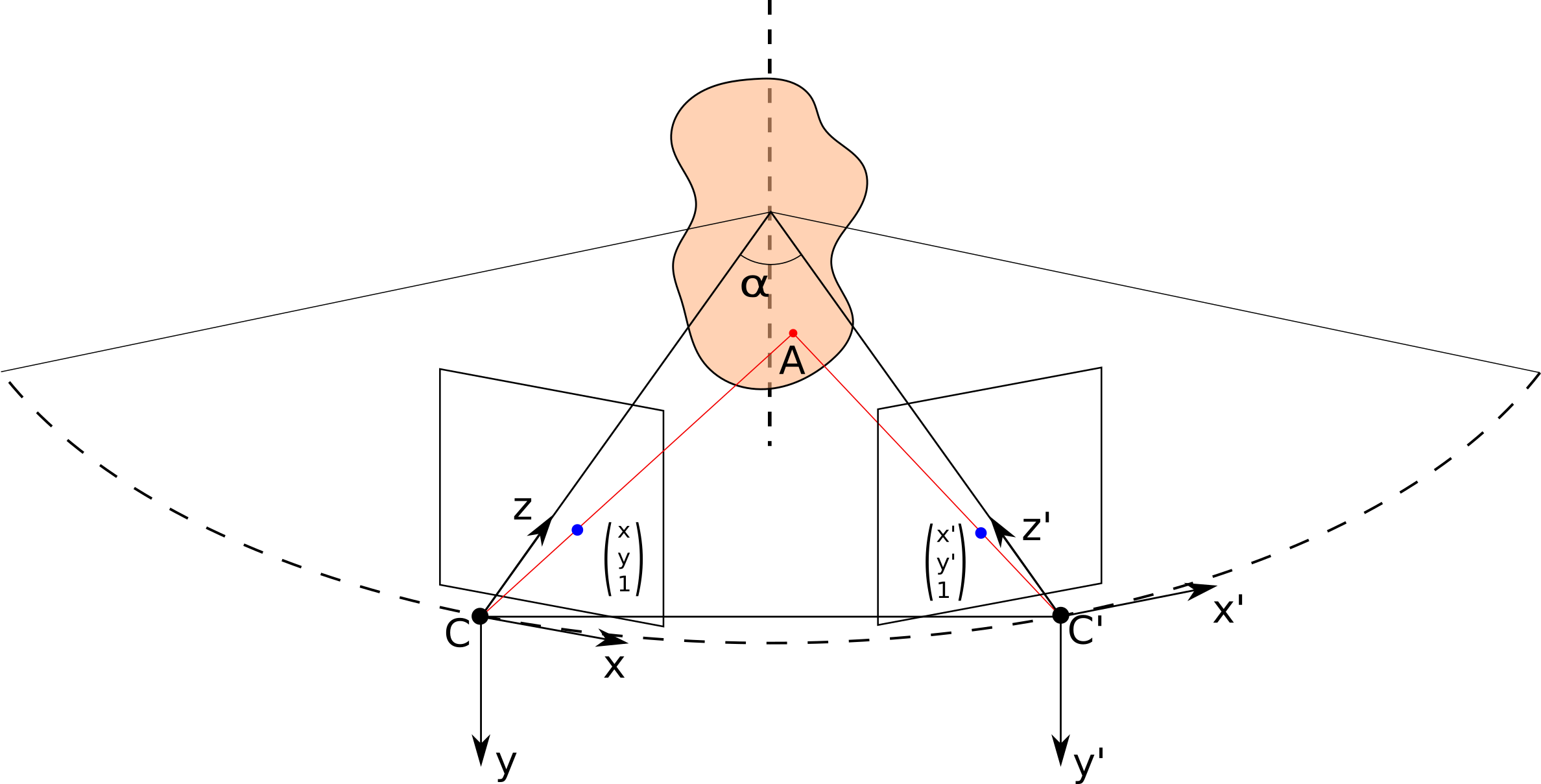}
    \caption{Circular camera motion}
    \label{fig:geometry}
\end{figure}

Consider a camera moving in a circle.
The optical axis of the camera lies in the horizontal plane, and the object itself is stationary (see Fig.~\ref{fig:geometry}).
The task is to estimate the angle of rotation of the camera between two positions from the images obtained in these positions.
It can be shown that to solve this problem it is enough to know the coordinates of the projection onto the image plane of one 3D point in each position.
Let $ q = (x, y, 1) ^ T, q '= (x', y ', 1) ^ T $ be the homogeneous coordinates of the same point for the first and second camera positions, respectively.
These coordinates are linked by an essential matrix:
\begin{equation}
    E = \big[t\big]_{\times} R 
    \label{eq:ess_mat}
\end{equation}
где $R$ -- rotation matrix around the $y$ axis, $\big[t\big]_{\times} = 
r\begin{pmatrix} 
0 & -2\sin^2\frac{\alpha}{2} & 0\\
2\sin^2\frac{\alpha}{2} & 0 & -\sin\alpha\\
0 & \sin\alpha & 0\\
\end{pmatrix}$ -- is the matrix representation of the cross product with translation vector $t$, $r$ -- radius of the circle along which the camera moves. The system has one degree of freedom. Therefore it is possible to parameterize the essential matrix with a single parameter, in our case the camera rotation angle:
\begin{equation}
    E = \begin{pmatrix} 
    0 & -2\sin^2\frac{\alpha}{2} & 0\\
    2\sin^2\frac{\alpha}{2} \cos{\alpha} - \sin^2\alpha & 0 & -2\sin^2\frac{\alpha}{2}\sin\alpha - \sin\alpha\cos\alpha \\
    0 & \sin\alpha & 0\\
\end{pmatrix}. 
\end{equation}

According to~\cite{hu2008epipolar}, the points $ q $, $ q' $ are connected by the relation $ q'^TEq = 0 $. Having expanded this expression, we obtain an expression for the camera rotation angle:
\begin{equation}
    \label{eq:angle}
    \alpha = 2 \arctan{\Big(\frac{y-y'}{x'y + xy'}\Big) }.
\end{equation}
The rotation matrix is uniquely derived from this angle and, if the radius of the circle is specified, the translation vector can be derived as well.

\section{Outliers removal}

To find point correspondences on a pair of tomographic projections, the algorithm for detecting and matching keypoints is used.
It is prone to the wrong matches.
We propose to use the RANSAC-based algorithm as an algorithm for estimating the angle of axial rotation that is robust to outliers (wrong correspondences).

At each iteration of the algorithm an arbitrary pair of corresponding points is selected.
Using coordinates of chosen points the angle of rotation is calculated according to the equation~(\ref{eq:angle}). Next, the Sampson distance is calculated~\cite{hu2008epipolar} for each remaining pair of points: 
\begin{equation}
    d(q, q') = \sqrt{\dfrac {\left(q'^TEq\right)^2}{\left(Eq\right)_x^2 + \left(Eq\right)_y^2 + \left(E^Tq'\right)_x^2 + \left(E^Tq'\right)_y^2}},
\end{equation}
where $ E $ is the essential matrix. A pair of points is considered to be inlier (correct correspondence) if the error does not exceed a given threshold. If the proportion of such points exceeds a predetermined value, the model is considered to be a valid consensus. 
For all probable models, using the least squares method~\cite{golub1968least}, an angle estimate is calculated to minimize the sum of the squared errors of the form $ q '^TEq $ for all satisfying the model points pairs. The result of the algorithm is the angle chosen among all valid models at which the minimum squared error is achieved.

The required number of iterations of the RANSAC algorithm can be calculated as:
\begin{equation}
    N \geq \frac{\log(1 - p)}{\log(1 - (1-\epsilon)^n)},
\end{equation}
where $ p $ is the probability for $ N $ iterations to select at least one data set without outliers, $ \epsilon $ is the fraction of outliers, $ n $ is the number of points necessary to build the tested model (in our case, $ n = 1 $).

\section{Experimental Results}

A series of experiments was conducted to test the proposed algorithm. First of all, the rotation angle estimation algorithm was tested on synthetic data. The experiment made it possible to study the influence of the fraction of outlier correspondences and the spatial accuracy of the keypoints detector on the quality of the rotation angle estimation. Experiments on a real sequence of tomographic projections made it possible to evaluate the effect of the nature of X-ray images on the accuracy of the obtained estimates. The following sections describe in detail the methodology of the experiments and the obtained results.

For a relative quality assessment, the proposed algorithm was compared with two alternatives. The first algorithm calculates the median of all estimated angles for all pairs of corresponding points. The second algorithm is based on the analysis of a histogram of angles computed for all correspondences (hereinafter we will refer to this algorithm as a \textit{histogram}). In this algorithm, the range of angles was divided into several intervals. For each pair of correspondences, the rotation angle was calculated using the equation~\ref{eq:angle}. For each histogram interval, the number of pairs of points defining the angle from the corresponding interval was stored. Next, we selected the interval in which the largest number of angle estimates fell. Based on correspondences that fall into this interval, an angle was calculated with the least squares method.

\subsection{Synthetic data experiments}

\begin{figure}
    \centering
    \null\hfill
    \begin{subfloat}[][]{
        \centering
        \includegraphics[width=0.4\textwidth]{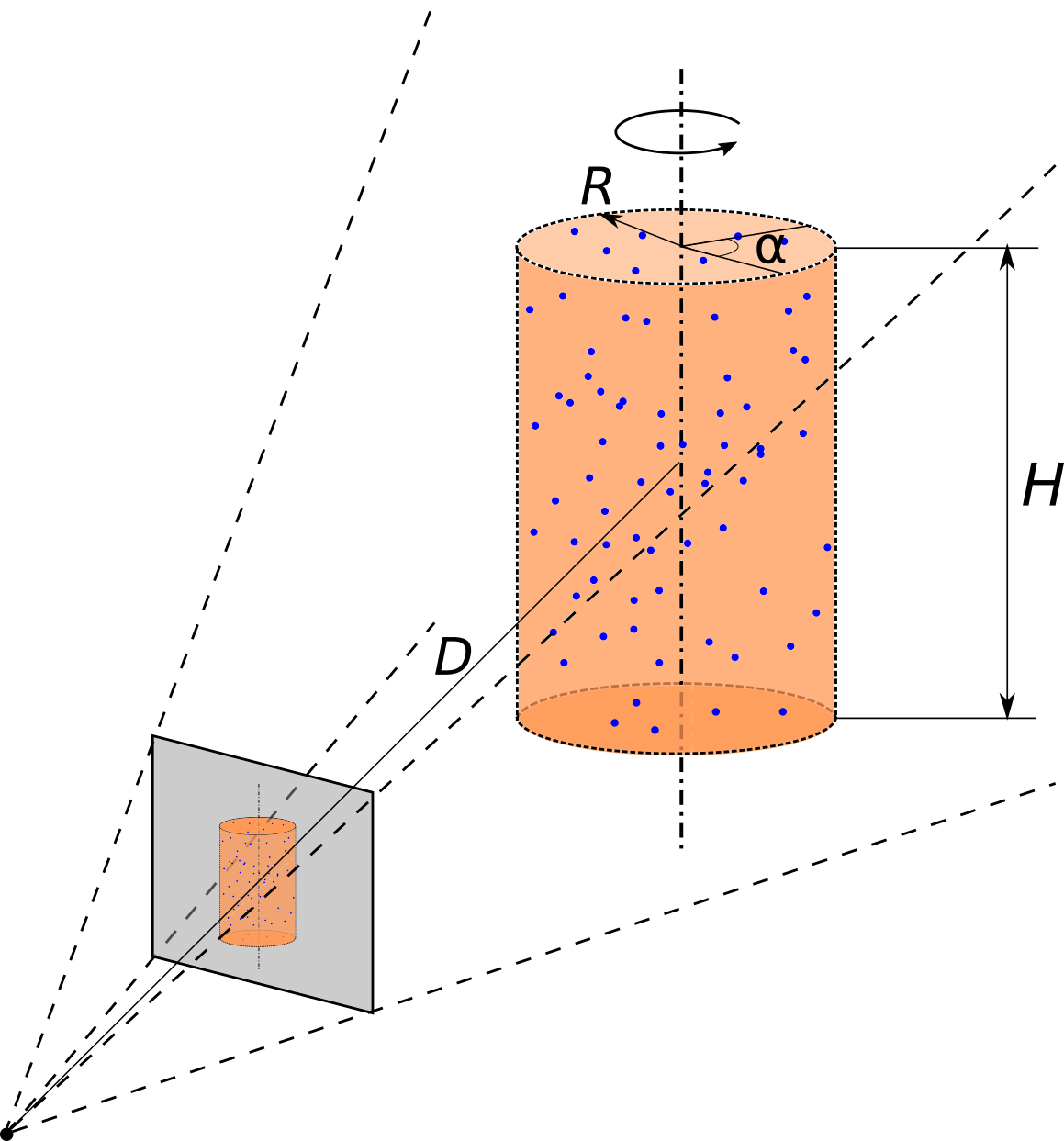}
        \label{fig:cylinder_gen}}
    \end{subfloat}
    \hfill
    \begin{subfloat}[][]{
        \centering
        \includegraphics[width=0.4\textwidth]{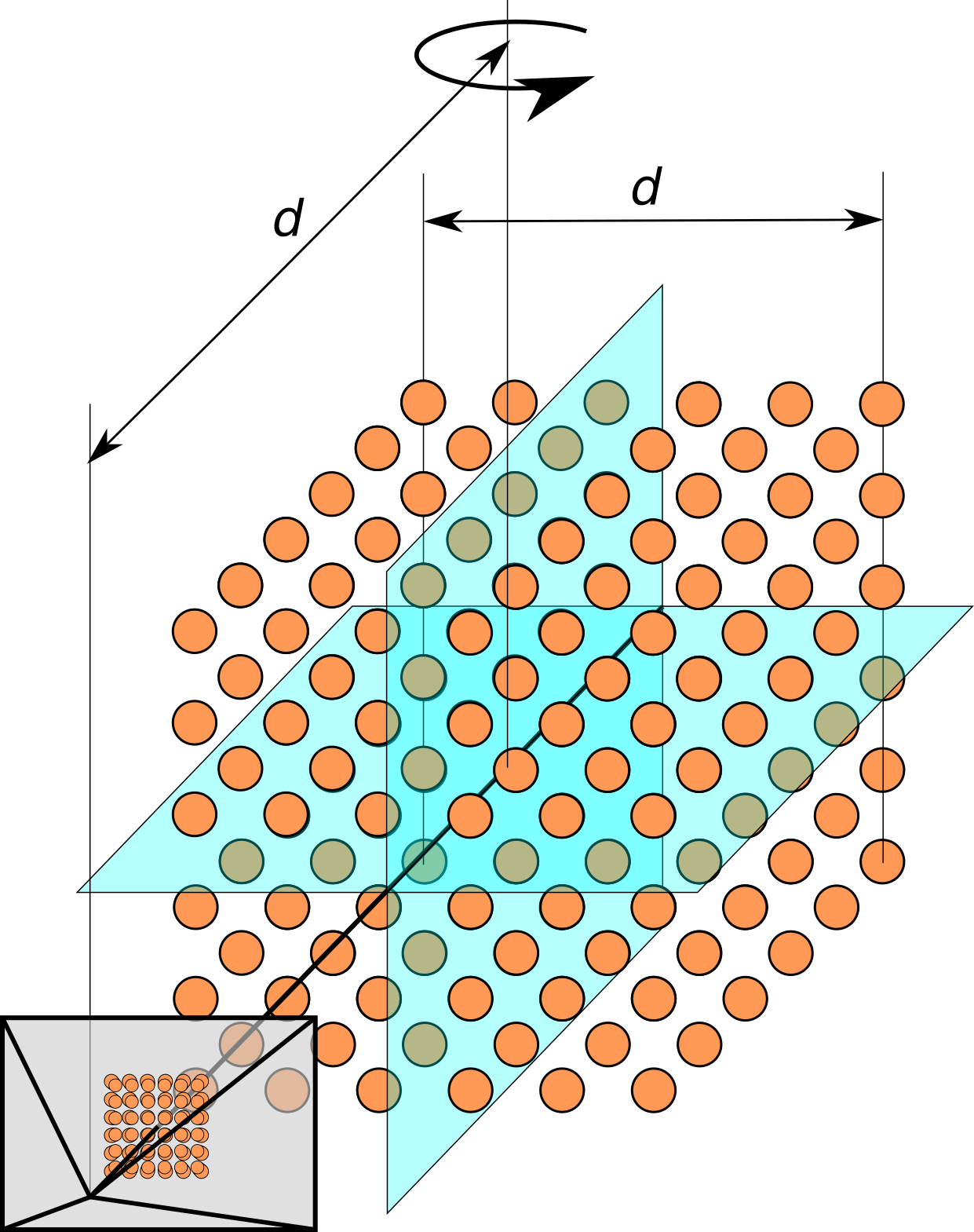}
        \label{fig:points_cube}}
    \end{subfloat}
    \null\hfill
    \hspace*{1.15in}
    \caption{(a): synthetic data generation scheme. (b): rotation angle equation conditioning experimental scheme.}
\end{figure}

In the experiment the following data generation procedure was used. Points were generated randomly (uniformly distributed) in the cylinder volume (see Fig.~\ref{fig:cylinder_gen} and Table~\ref{tab:gen_ransac}). The resulting point cloud was rotated by a predefined angle. For both sets of points (original and rotated), a transformation was made from the three-dimensional coordinates of the points to homogeneous coordinates. Then normal noise with a zero mean was added to both homogeneous coordinates of original and rotated points. Thus, a set of pairs of points in uniform coordinates was generated. Outliers were added to the resulting set of point pairs by randomly matching points before and after rotation of the point cloud.

\begin{figure}
    \centering
    \begin{subfloat}[][]{
        \includegraphics[width=0.65\textwidth]{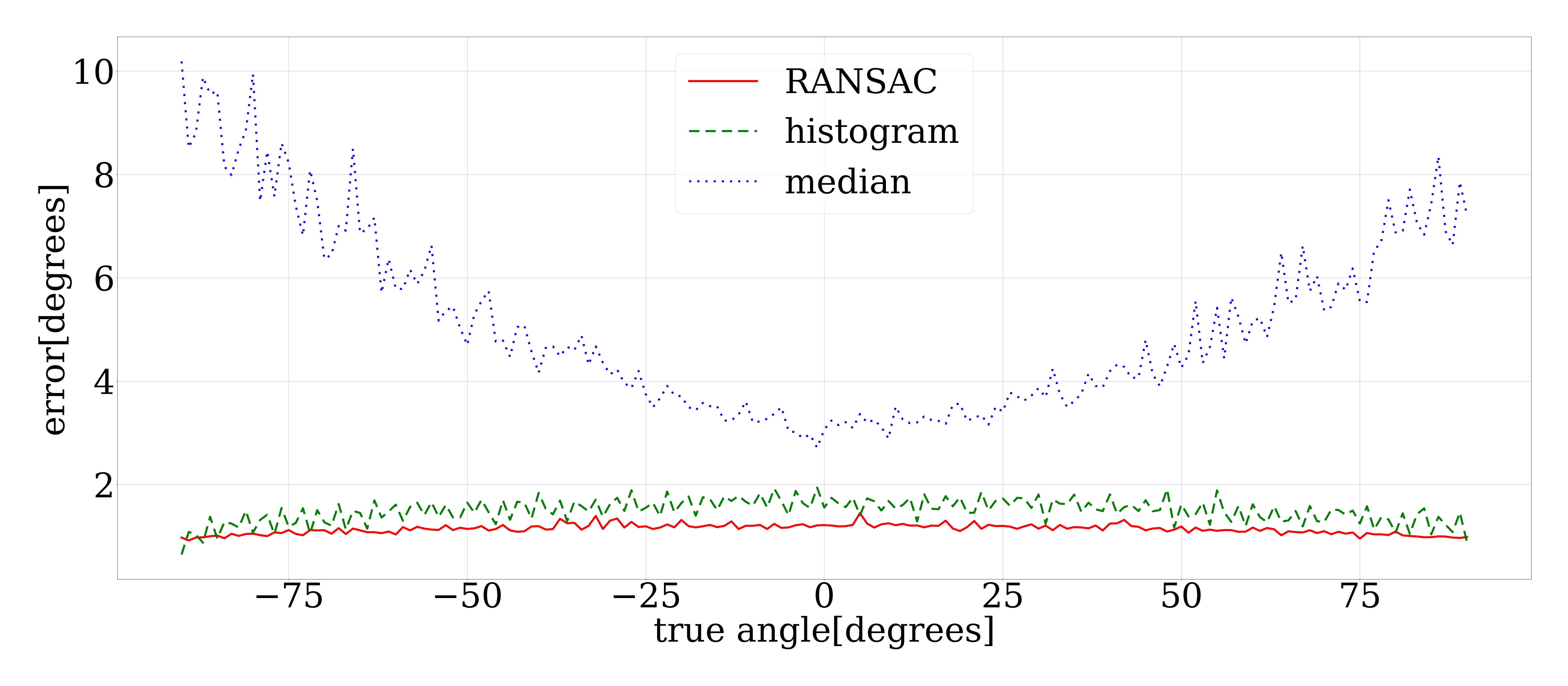}
        \label{fig:gen_comparison}}
    \end{subfloat}
    \begin{subfloat}[][]{
        \includegraphics[width=0.3\textwidth]{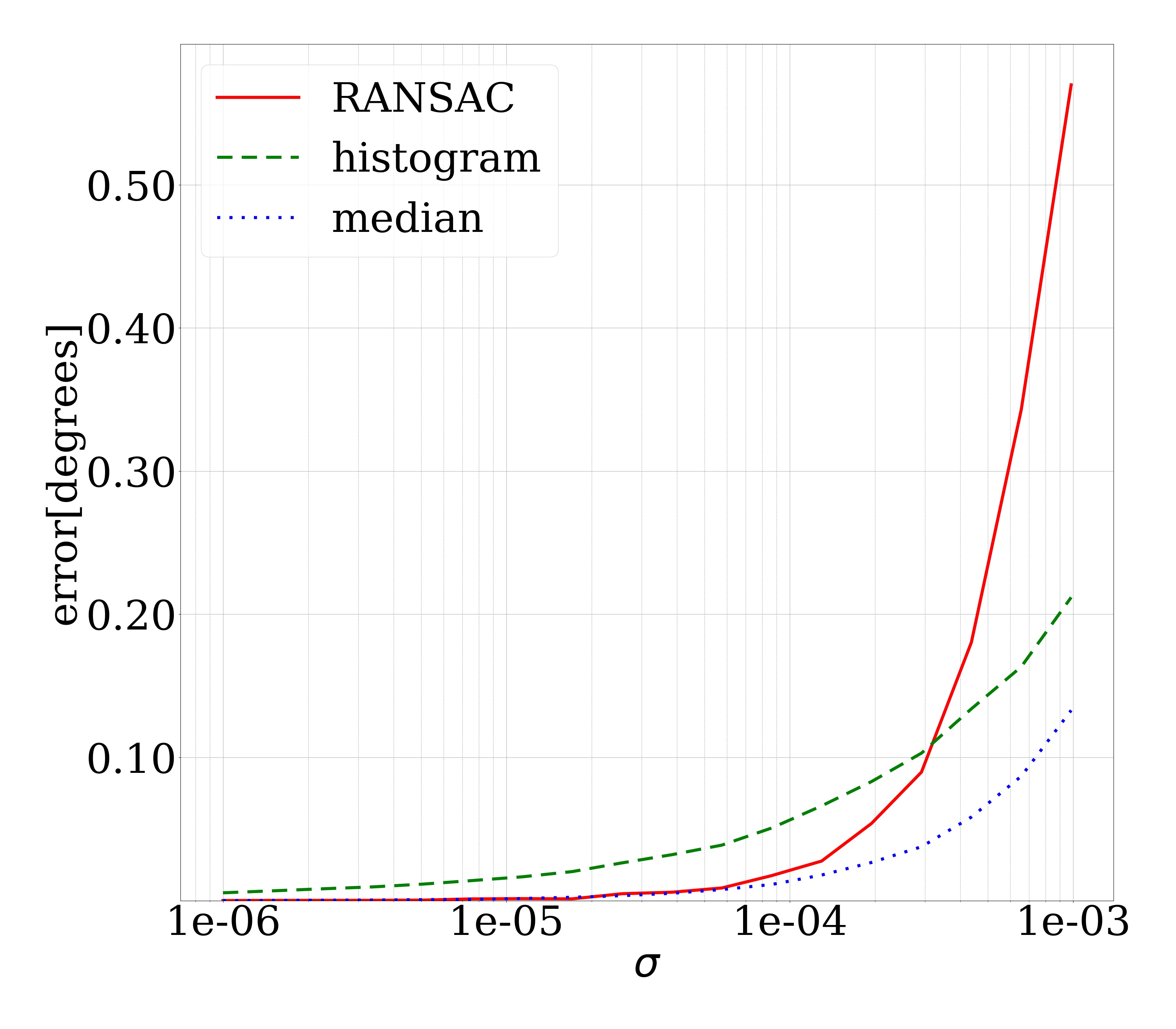}
        \label{fig:noise_test}}
    \end{subfloat}
    \caption{Mean absolute error dependency on the (a): rotation angle; (b): keypoints spatial detection noise.}
    \hspace*{1.15in}
\end{figure}

For rotation angles in the range of $ (- 90^{\circ}; 90^{\circ}) $ pairs of ьфесрув points were generated in the described above way. Each of the three described algorithms was tested on the whole range of angles. The angle estimation was repeated  $ 300 $ times for each angle of rotation. The mean absolute error of rotation angle estimation was calculated for each algorithm for each angle form the range. The RANSAC algorithm parameters are presented in the Table~\ref{tab:gen_ransac}. The experimental results are shown in Fig. ~\ref{fig:gen_comparison}.

The obtained results demonstrate that the proposed algorithm is superior in quality to the histogram and the median ones for the entire range of the rotation angles. This is due to the resistance of RANSAC to outliers, which inevitably arise when matching keypoints.

In addition, the influence of spatial keypoints detection noise on the rotation angle estimation accuracy was evaluated.
In this experiment, the same point generation parameters were used as those specified in the Table~\ref{tab:gen_ransac} except for the number of correct correspondences and outliers that were changed to $ 100 $ and $ 0 $, respectively.
The noise of keypoints detection was modeled by a normal distribution with zero mean.
The standard deviation varied in the range $ \left(10^{-6}; 10^{-3}\right) $. For each noise value, the measurement was carried out $ 1000 $ times and then the obtained error values were averaged. The error magnitude dependence on the noise standard deviation is shown in Fig.~\ref{fig:noise_test}.

It can be seen from the obtained results that with a sufficiently large noise, the error in the RANSAC estimates starts to exceed the errors of other algorithms. First of all, this is due to the fact that during the experiment the Sampson distance threshold did not change, as a result of which correct point correspondences were considered as outliers.

\begin{table}
    \centering
    \caption {Synthetic data experiment parameters.}
    \begin{tabular}{|l|c|}
        \hline
        \hline
        \textbf{Parameter}                              & \textbf{Value} \\ \hline
        Distance to cylinder axis, $D$                      & 200               \\ \hline
        Cylinder height, $H$                                & 230               \\ \hline
        Cylinder radius, $R$                                & 115               \\ \hline
        Keypoints detection noise model         & $\mathcal{N}(0,\,\sigma^{2})$        \\ \hline
        Standard deviation of detection noise, $\sigma$                         & 0.0001            \\ \hline
        Number of outliers                                  & 70                \\ \hline
        Number of inliers (correct correspondences)                 & 30                \\ \hline
        Probability to choose the model without outliers within N iterations  & 0.999     \\ \hline
        Requred fraction of inliers to consider the consensus valid & 0.25              \\ \hline
        Sampson distance threshold                        & 0.01            \\ \hline \hline
    \end{tabular}
    \label{tab:gen_ransac}
\end{table}

\subsection{Experiments on tomographic projections}

\begin{figure}[h]
    \centering
        \includegraphics[width=0.8\textwidth]{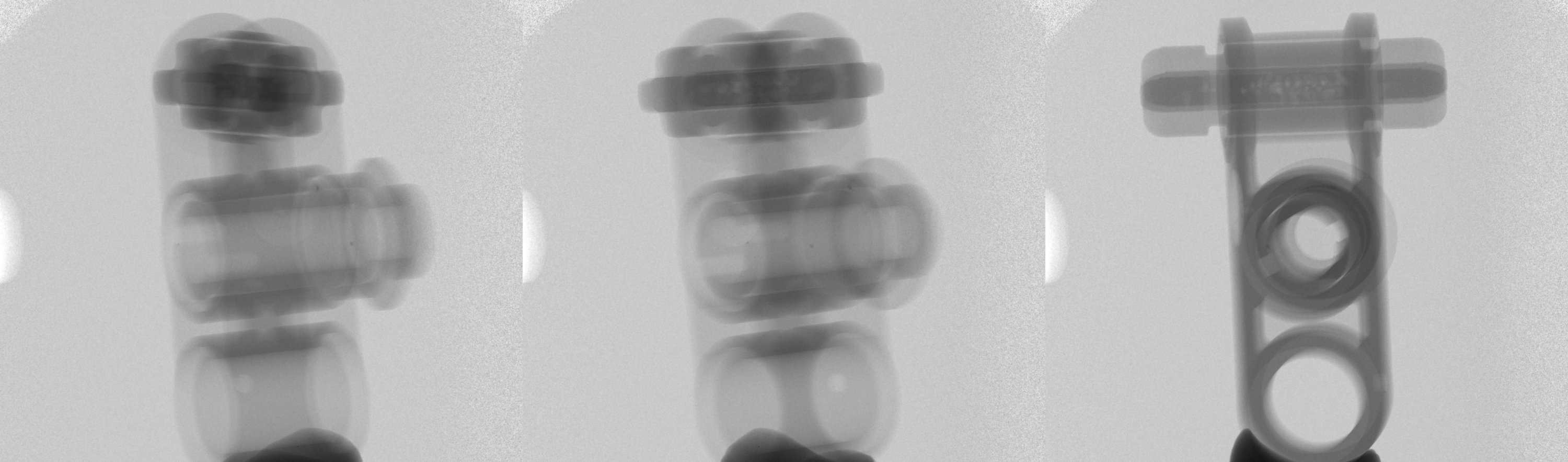}
    \caption{Samples of tomographic projections.}
    \label{fig:dataset}
\end{figure}

To test the proposed algorithm, a set of tomographic projections was collected at the X-ray microtomograph developed and operating at the Federal Research Center for Crystallography and Photonics of Russian Academy of Sciences~\cite{buzmakov2018laboratory, buzmakov2019laboratory, chekanov2019xray}.
Projections of a rotating object were formed with a step of $ 0.5 $ degrees in the range of $ 0 - 200 $ degrees (see Fig.~\ref{fig:dataset}).

The SURF (Speeded Up Robust Features) algorithm for keypoints detection and descriptions was used to obtain point correspondences on a pair of tomographic projections. The Hessian threshold was set to $ 100 $. For a detailed description of the algorithm and its parameters the reader can refer to the original paper~\cite{bay2006surf}.

The best accuracy was achieved with the following parameters of the RANSAC algorithm: the probability to choose as a model in $N$ iterations a pair of points that is not an outlier is $ 0.999 $; fraction of correct correspondences -- $ 0.05 $; the Sampson distance threshold at which the corresponding points are still considered as valid correspondence -- $ 8 \cdot 10^{-4} $; the minimum fraction of inliers needed to considered a consensus valid -- $ 0.6 $. A total of 398 pairs of tomographic projections were tested. The true value of the axial rotation angle for each pair was $ 1 $ degree. The Table~\ref{tab:real_dat_perf} presents the results of the experiment.

\begin{table}
 \centering
\caption {Real data experimental results}
\label{tab:real_dat_perf}
\begin{tabular}{|p{0.5\linewidth}|c|c|c|}
\cline{2-4}
\multicolumn{1}{c|}{} & RANSAC & Histogram & Median\\ \hline
Number of projections for which the rotation angle was successfully estimated & 341 & 396 & 398  \\ \hline
Mean absolute error & 1.45& 1.64 & 2.85\\ \hline
Standard deviation & 2.59 & 5.41  & 9.50\\ \hline
Minimal error   & 0.0051  & 0.0016 & 0.0092\\ \hline
Q1 (25\%)  & 0.52            & 0.24     & 0.47         \\ \hline
Q2 (50\%)  & 0.86            & 0.60     & 0.93          \\ \hline
Q3 (75\%)  & 1.49            & 1.98     & 1.81         \\ \hline
Maximum error & 33.24 & 72.12 & 106.64 \\ \hline \hline
\end{tabular}
\end{table}

It can be seen from the results that RANSAC could not estimate the angle on 47 image pairs. In two cases the histogram algorithm did not have a peak above one pair per interval, and, therefore, could not estimate the angle. On average the error of the proposed algorithm is less than that of the other two. The maximum error value of the histogram method is $ 72 $ degrees, for RANSAC this value is half as much: $ 33 $ degrees. The median shows the worst results. All three methods showed an average error comparable to the true angle of rotation. Such a big error can be explained by the fact that when rotating $ 1 $ degree coordinates of object points undergo a small change. For instance, when rotating $ 1 $ degree, the change in the $ y $  coordinates (see the equation~\ref{eq:angle}) of the corresponding points is approximately $ 1 $ - $ 3 $ pixels, which is comparable to the keypoints detection spatial noise. An increase in the rotation angle between the projections would positively influence the results, however impossible due to the inability of the SURF algorithm to match points of tomographic projections at rotation angles exceeding $ 3 $ degrees for the data set used in this work. Thus, a reliable point matching of tomographic projections for large rotation angles seems to be a further direction of work. In addition, the distance from the optical axis and the axis of rotation of the object affects the magnitude of the point coordinates change. Several additional experiments were carried out to to test if the problem of rotation angle estimation is well-conditioned depending on these factors.

\subsection{Conditioning of rotation angle estimation problem on point coordinates}

\begin{figure}
    \centering
    \includegraphics[width=0.7\textwidth]{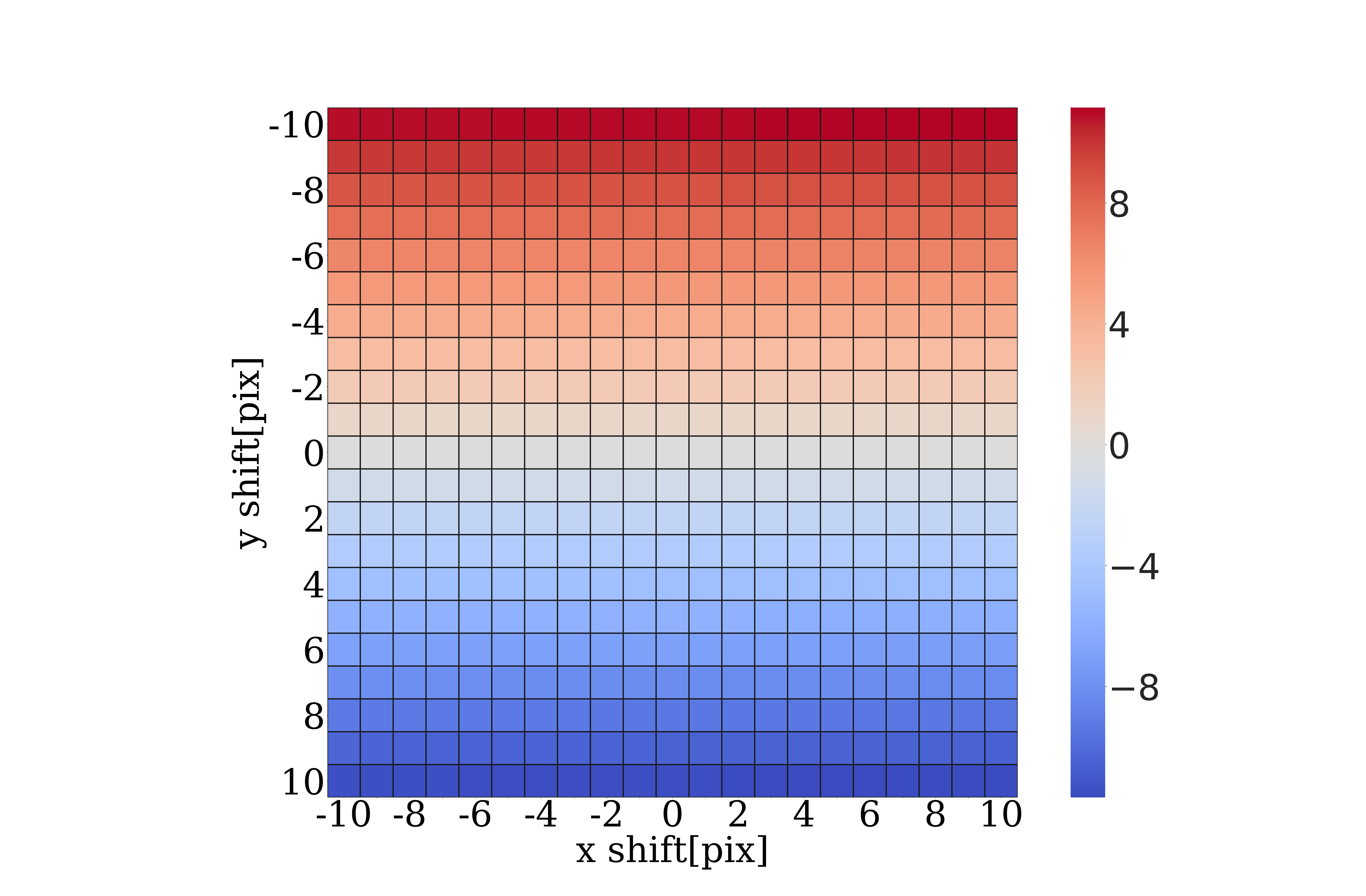}
    \caption{Rotation angle estimation error depending on the shift of one point of the pair. Axes denote shift of the point in image plane in pixels.}
    \label{fig:calc_angles}
\end{figure}

We have tested the robustness of the rotation angle estimation made by proposed RANSAC-based algorithm to spatial keypoints detection error.
For a pair of real tomographic projections among all the correspondences selected as a valid consensus, the best one was selected (the one that has minimal sum of squared errors). 
Further, in the best pair, the position of the first point was fixed and the position of the second one was changing; the rotation angle estimation was re-calculated for each shift of the second point. 
The true value of the axial rotation angle was 1 degree. 
The calculated angle values are shown in Fig.~\ref{fig:calc_angles}. 
It can be seen that shift along the $ x $ axis have little effect on the estimation, while an error of $ 1 $ pixel by $ y $ leads to a change in the result by more than a degree. In the experiments on tomographic projections the average point shift along the $ y $ axis caused by rotational motion was equal to $ 1.75 $ pixel.

\begin{figure}[t]
    \centering
    \null\hfill
    \begin{subfloat}[][]{
        \centering
        \includegraphics[width=0.3\textwidth]{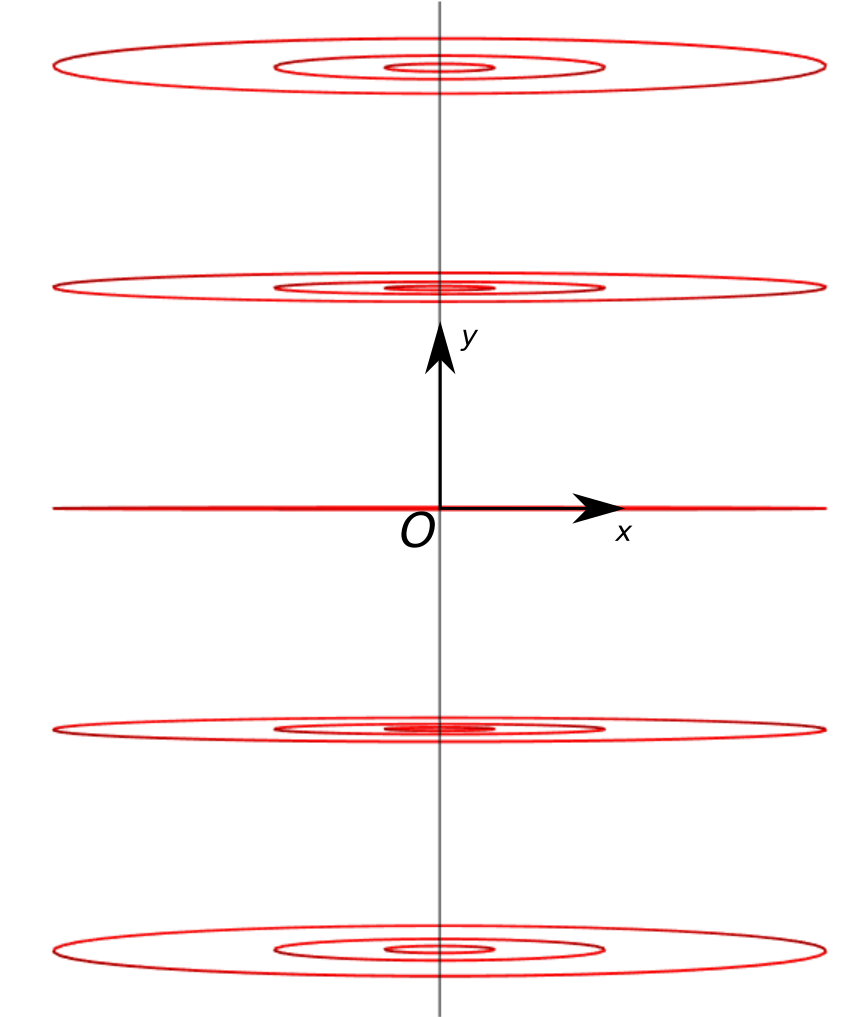}
        \label{fig:circles}}
    \end{subfloat}
    \hfill
    \begin{subfloat}[][]{
        \centering
        \includegraphics[width=0.6\textwidth]{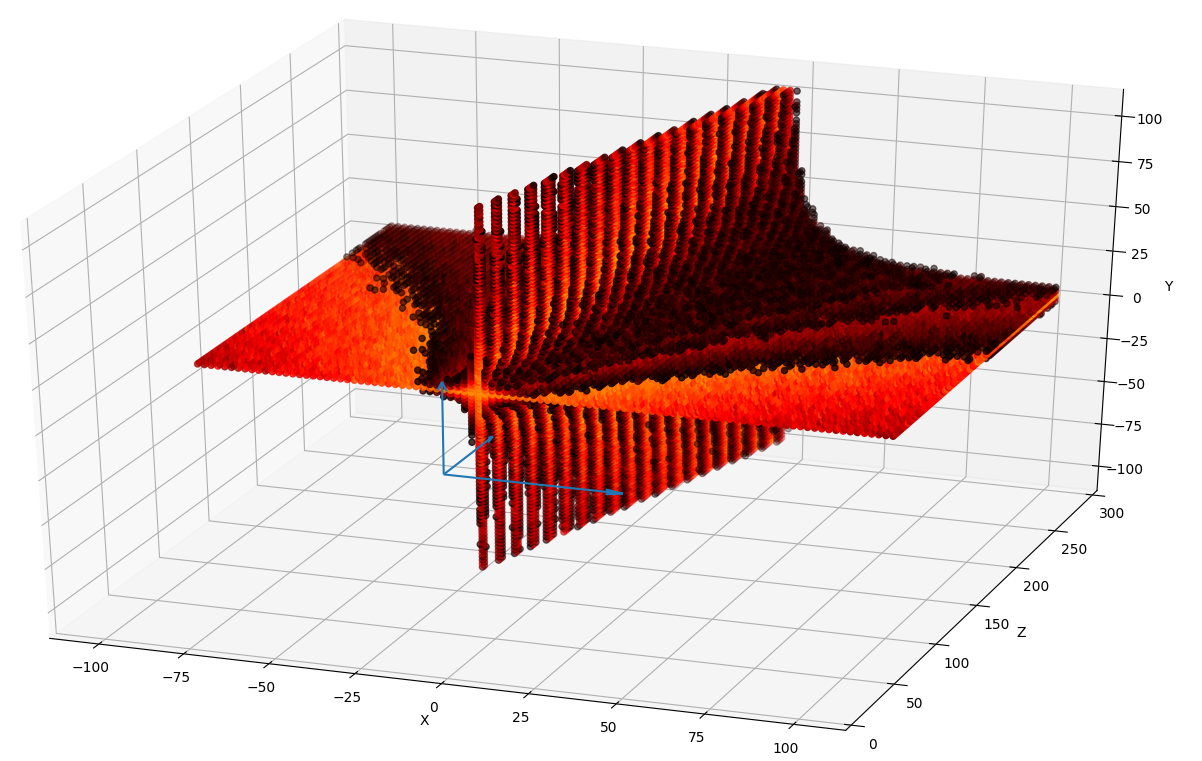}
        \label{fig:error_map}}
    \end{subfloat}
    \null\hfill
    \hspace*{1.15in}
    \caption{(a): object points trajectories projected into the image. O -- principal point. (b): rotation angle estimation error distribution for a 3D cubic lattice. The brighter color indicates a larger error.}
\end{figure}

As was mentioned in the previous section, there are possible points locations for which the equation for calculating the rotation angle is ill-conditioned. 
Fig.~\ref{fig:circles} illustrates the trajectories of some points of the rotating object. It can be seen that the ellipses in the vicinity of the horizontal plane passing through the camera optical center have a smaller minor axis, than the ones on the distant planes. 
Proximity to optical center leads to a larger error in calculating the angle due to the smaller signal-to-noise ratio while determining the vertical component of the point motion with keypoints detection and matching. To study how the equation~\ref{eq:angle} is conditioned on the position of an object's point in space, the following experiment was conducted. The cubic lattice of size $200$ was generated with the center at a distance of $200$ from the camera center (see Fig.~\ref{fig:points_cube}). Then it was rotated by a arbitrary chosen angle of $ 21 $ degrees. Then, similarly to the experiment on synthetic data, the coordinates of the points were transformed to homogeneous with the addition of normal noise ($ \sigma = 0.004 $). Based on each pair of points (original and rotated), the rotation angle and its deviation from the true value were estimated. The result was averaged over $100$ measurements. To visualize the results, points whose average error was less than $ 60 $ degrees were discarded. This step was performed to highlight regions with the highest error. The result of the experiment is shown in Fig.~\ref{fig:error_map}. The remaining points (points leading to the largest error value) lie in two planes. One of them is horizontal and intersects the principal point when projected on the image plane. Another plane is passing through the axis of rotation at an angle equal to the half of rotation angle ($ 10.5 $ degrees in our experiment). When substituting the points lying in the described areas into the equation~\ref{eq:angle}, the uncertainty of the form $ \frac{0}{0} $ is obtained. So the problem of rotation angle estimation is ill-conditioned for points lying in these areas. 

Based on the experimental results it was estimated that the detected keypoint must lie on the distance of $0.9$ from the optical center to obtain an angle estimation error less than $0.5$ degrees. For the presented datset the furthest detected object point lies on the distance of $0.3$ from the optical center which leads to the $1$ degree error (see Table~\ref{fig:dataset}).

\section{Conclusion}

A RANSAC-based algorithm for axial rotation angle estimation of an object from its tomographic projections is proposed. The proposed algorithm requires only one correct correspondence of the points of two projections. An experimental comparison of the proposed algorithm with baseline algorithms on synthetic and real data is conducted. It is shown that the proposed algorithm is superior. Also, a series of experiments was conducted aimed at studying the conditionality of the problem of rotation angle estimation from a pair of corresponding images points. So, on the one hand, it is shown that at small angles of rotation, the spatial noise of keypoints detector can be comparable to a change in the point coordinates caused by rotation, which means that keypoints detection and matching algorithm is needed that allow for reliable match of tomographic projections points at large angles of rotation of the object. On the other hand, the obtained results demonstrate that the proximity of the object’s point to the axis of rotation and the horizontal plane crossing the camera optical axis makes the problem ill-conditioned and in the extreme case makes estimation uncertain. Thus, when choosing points, it is necessary to filter the points lying in the mentioned areas. It is important to note that in tomography, the peripheral areas of an object are subject to less overlapping signals coming from different depths of the object, which means that potentially such areas are better described by keypoint algorithms developed for images formed in an opaque world model. Thus, the areas of the object far from the optical axis of the camera and the axis of rotation of the object itself are of the greatest interest for estimating the magnitude of axial rotation.

It was shown that to apply the proposed algorithm for solving practical problems, it is necessary to conduct a preliminary analysis of the conditionality of the problem for the particular object and the keypoints detected on it. The proposed algorithm can be further improved by developing mechanisms for keypoints detection and description specifically optimized for digital radiographic images. Another direction of development is to minimize the error in determining the angle of axial rotation when comparing different pairs of the sequence of images, not just neighboring ones.

\acknowledgements
The reported study was partially funded by RFBR, project number 18-29-26037.

\bibliographystyle{unsrt}
\bibliography{bibliography}


\clearpage

\section*{AUTHORS' BACKGROUND}

\begin{table}[h]
	\centering
	\begin{tabular}{ | c | c | p{45mm} | p{60mm} | } \hline
		Name & Title & \centering Research Field & Personal website \\
		\hline
		Mikhail Chekanov & Undergraduate student & Computer vision & \texttt{https://www.researchgate.net/\newline profile/Mikhail\_Chekanov}\\
		\hline
		Oleg Shipitko & PhD Student & Robotics, computer vision & \texttt{https://www.researchgate.net/\newline profile/Oleg\_Shipitko}\\
		\hline
		Egor Ershov & PhD  & Computer vision, pattern recognition, image processing & \texttt{https://www.researchgate.net/\newline profile/Egor\_Ershov3} \\
		\hline
	\end{tabular}
\end{table}

\end{document}